# Learning Deep Representations for Word Spotting Under Weak Supervision

Neha Gurjar, Sebastian Sudholt, Gernot A. Fink
Department of Computer Science
TU Dortmund University
44221 Dortmund, Germany
Email: {neha.gurjar, sebastian.sudholt, gernot.fink}@tu-dortmund.de

*Abstract*—Convolutional Neural Networks have made their mark in various fields of computer vision in recent years. They have achieved state-of-the-art performance in the field of document analysis as well. However, CNNs require a large amount of annotated training data and, hence, great manual effort. In our approach, we introduce a method to drastically reduce the manual annotation effort while retaining the high performance of a CNN for word spotting in handwritten documents. The model is learned with weak supervision using a combination of synthetically generated training data and a small subset of the training partition of the handwritten data set. We show that the network achieves results highly competitive to the state-of-the-art in word spotting with shorter training times and a fraction of the annotation effort.

## I. INTRODUCTION

Convolutional Neural Networks (CNN) achieve remarkable results across a variety of different computer vision domains. Most recently, they were also successfully introduced to the field of document image analysis and word spotting in particular. The task in word spotting is to retrieve relevant parts of a document image collection with respect to a query. Most often, the query is either a document image itself (Query-by-Example, QbE) or a textual representation (Query-by-String, QbS).

The exceptional performance of CNNs often times comes at the cost of having to supply a large amount of diverse, annotated training samples. This is, of course, a characteristic that is highly undesirable for word spotting systems. When supplied with a new document image collection, this requires the user to first perform tedious labeling work before being able to perform word spotting. In some instances, this might already solve the task of finding specific word images, which would render the entire word spotting approach moot. In an ideal scenario, a word spotting system requires little to no data which had to be manually annotated by the user. However, this is, of course, detrimental for the overall performance of the CNN. If the annotated data at hand is only available in small quantities, a common approach for other tasks is to finetune a CNN previously trained on a larger data set. This of course requires that the larger data set stems from the same domain as the smaller data set that the CNN is supposed to be finetuned to. For example, it can be expected that finetuning a CNN trained for object detection to a word spotting benchmark does not yield a competitive performance.

The finetuning approach, however, does not mitigate annotation efforts but rather shifts them: The larger data set still had to be annotated at some point in time. It would be desirable here to be able to generate an annotation for training with as little human effort as possible.

In this work, we investigate the effects of training a CNN for word spotting under weak supervision. Here, we define weak supervision as having little manually annotated data and have the CNN trained mainly on synthetically generated data. The questions we focus on are: Can good word spotting results be achieved by training a CNN on purely synthetically generated data for both contemporary and historical documents? If not, what is an adequate amount of word images that have to be manually annotated to achieve competitive performance?

For evaluation purposes, we make use of the recently introduced PHOCNet [1]. We conduct experiments for both historic and contemporary document images.

First, we evaluate using purely synthetic images rendered from Latin script fonts for training a CNN. Only the mapping between textual character and visual character in the font has to be established by a human and word images can be generated unsupervisedly from there on.

Second, we evaluate using an increasing number of training images from the respective data sets in conjunction with the synthetic images. We define this approach as weak supervision as the number of training images to be annotated by a human is limited to a small amount.

We show empirically, that using 250 annotated word images (roughly one page of a historical document) in conjunction with a number of synthetically generated images is enough for a deep CNN like the PHOCNet to be able to achieve competitive results.

## II. RELATED WORK

### A. Word Spotting

Key Word Spotting or simply word spotting originally stems from the field of audio and speech processing where it was used to detect certain key words in an audio stream. Different works focused on transferring the success of word spotting to the handwritten domain. One of the earliest works in this regard made use of XOR-maps and Euclidean Distance Mapping in order to spot words in binarized handwritten

images [2]. Ensuing approaches often times focused on challenging historic documents. Here, methods were applied which had proven successful in handwriting recognition. Especially sequential models such as Dynamic Time Warping [3], Hidden Markov Models (HMM) [4], [5] and Long Short Term Memory networks (LSTM) [6] were successfully applied to word spotting.

Apart from the sequential models, methods from other fields of computer vision found their way into word spotting as well. Especially local image descriptors were applied to word spotting problems quite frequently in the form of SIFT, e.g. [7]–[9], or HOG-based descriptors [10]. While most often local descriptors are pooled into either Bag-of-Features or Spatial Pyramid representations, Bag-of-Feature HMMs [11], [12] successfully combine local descriptors and sequential models for word spotting.

The recent success of deep *Convolutional Neural Networks (CNN)* in other fields of computer vision made researchers investigate their use for word spotting as well. Combining CNNs with attribute representations such as the *Pyramidal Histogram of Characters (PHOC)* [13] achieves the current state of the art in a number of word spotting applications [1], [14]–[16].

## B. Synthetic Training Data

In contrast to our focus of reducing manual annotation efforts, early works on using synthetic data in document image analysis were primarily motivated by increasing the training data size of already existing data sets. This was meant to improve classification performance in handwriting recognition tasks [17], [18] or to supply data sets for new script types for which obtaining document images and annotations is difficult [19]. A detailed overview of different approaches for synthesizing images from online-handwritten trajectories as well as offline images is given in [20].

A straight forward approach for synthesizing training data for a classifier is to use machine printed text. However, this approach only yields competitive performance if the printed and handwritten script are similar. For example, Arabic machine printed and handwritten characters are both of cursive nature which allows for training a model from synthetically generated training data. This concept was successfully demonstrated in [21]. Here, the authors create a training set from various rendered Arabic fonts. This training set is then used to train an HMM which is adapted to handwritten text at test time in an unsupervised manner.

In order to allow a classifier to be trained on Latin script from machine printed text, [22] proposes the HW-SYNTH data set. It features a large number of word images rendered from handwritten-like Latin fonts. This very data set is used in [14] in order to pre-train a CNN for word spotting. The resulting CNN is then finetuned on other data sets. Word images are then represented by deep features extracted from the CNN. This approach differs from ours as it attempts to increase the training data, similar to [17] and [18], instead of aiming at reducing manual labeling costs.

## III. METHOD

### A. Learning With Weak Supervision

Unsupervised learning in the context of machine learning implies that a method does not require any annotated training data. Generating a Bag-of-Features representation [8], [9], [13] is an example of unsupervised learning of features. On the other hand, with supervised learning, a model is trained with the help of labeled training data which typically has to be supplied by a human. In the context of word spotting, we define weak supervision as training with minimal manual annotation effort. The training process may still require annotated data, however most training labels are generated without any human interaction, e.g., by creating synthetic data for which the labels are already available. A small amount of annotated training data may still be required in order for a system to achieve a desirable performance, which still has to be supplied by a human annotator. However, the amount of data to be annotated and the effort for computation is limited substantially compared to supervised learning where all training data has to be annotated. As a result, the human effort and time taken in order to prepare the training data is, also, reduced to a large extent.

The main problem with the weakly supervised approach using synthetic data is the variability in real handwritten text compared to synthetically generated text. This variability can be classified into two types, namely, inter-class and intra-class variation. Intra-class variation, in the context of word spotting, may include the difference in the visual appearance of the same word written by multiple authors due to different writing styles. Unsupervised techniques are unable to differentiate between these two types of variations of a data set. On the other hand, supervised techniques are able to identify different classes because of the annotated training labels. Hence, they are inherently able to distinguish between inter-class and intra-class variation of the given data set. CNNs are quite capable of adapting to a large intra-class variation or differentiating between classes despite a small inter-class variation in the data set. However, in order to differentiate between these two types of variation, the CNN needs to be presented with a sufficient amount of examples.

In our approach, we introduce a system which attempts to retain the ability of the supervised techniques to adapt to variation in text while keeping the amount of training data low, and hence reducing the annotation effort. For this, we make use of the HW-SYNTH data set [14] in order to train a CNN with a vast amount of handwritten-like images. We expect the vast amount of fonts used in this data set to represent possible variabilities in handwritten text to a reasonable degree. As CNN, we chose the recently published PHOCNet [1] as it achieves state-of-the-art results in word spotting and can be trained in an end-to-end fashion. We then perform word spotting as is usually done with the PHOCNet by computing attribute representations for a given word image (in this case the PHOC) and performing retrieval through a nearest neighbor search in the attribute space. For our system, no manual annotation of words is required. However, it must be noted that this is not an unsupervised approach in the context of word spotting, since for synthetically generating the word images, a mapping between machine readable characters and rendered handwritten-like characters has to exist. We evaluate the performance of this trained model using handwritten data sets. While our approach is capable of determining the specific attribute representations for real handwritten words to some extent, it faces limitations due to the difference in styles between the synthetically generated and handwritten word images as can be seen from the results.

In order to improve the performance of the model trained only on the synthetic data, we evaluate a second approach, namely finetuning the PHOCNet trained on synthetic data to small amounts of real handwritten data. This allows the CNN to adapt to the characteristics of the data set with minimal annotation effort. A major advantage of finetuning is that the amount of in-domain training data used to finetune a pre-trained network is much smaller as compared to what would, otherwise, be required to train the network from scratch.

It should be noted that the finetuning approach bares similarities to the deep features approach presented in [14]. However, there is a distinct difference between the two which is important to highlight: The deep features approach uses *the synthetic data in addition to the annotated training samples* and is simply there to enlarge the amount of data available for training. In contrast, our approach uses *few annotated samples in addition to the synthetic data*. We do not focus on enlarging the training samples but rather reduce it as we wish to only use as many annotated training samples as are absolutely necessary in order to achieve competitive performance. This way, we can evaluate how many annotated training samples are really necessary for the PHOCNet if synthetic data such as the HW-SYNTH data set is already available. We can thus empirically find a lower boundary for which we expect the PHOCNet to perform well on new data sets, too.

### B. PHOCNet Architecture

In this section, we briefly revisit important aspects of the CNN used in this work, namely the PHOCNet [1]. This neural networks achieves state of the art results in a number of word spotting benchmarks [16]. It does so by predicting an attribute representation for a given word, namely the *Pyramidal Histogram of Characters (PHOC)*. This attribute representation is then used in order to perform word spotting. Effectively, it replaces the AttributeSVM in the embedded attribute framework presented in [13]. While the PHOCNet's architecture is similar to the VGGnet [23], it exhibits a number of design choices with the aim of performing well specifically on word images. The first such choice is the use of a Spatial Pyramid Pooling layer after the convolutional part. First introduced in [24], it enables the network to be given an input of varying size while still producing a representation of constant size which can then be forwarded to the ensuing fully connected layers. The output of the SPP layer with max-pooling consists of the maxima taken from local spatial bins of the last feature map. The entire feature map is divided into an increasing number of bins in subsequent steps for a fixed number of steps. Since the bins have sizes proportional to the input feature map, the number of bins, and hence the size of the output of the SPP layer, is fixed despite a varying size of input images.

Another important aspect is the output layer of the PHOCNet and its activation function. Training labels for CNNs in recognition tasks are typically 1-hot vectors signifying the target class of the input. However, the PHOCNet is tasked with predicting an attribute representation where multiple elements in can be 1. Hence, the last layer of the PHOCNet uses the sigmoid instead of the ubiquitous softmax as activation function in the last fully connected layer. Training the PHOCNet with the binary cross entropy loss allows for predicting an attribute representation.

The PHOCNet employs a number of regularization techniques in order to avoid overfitting. A dropout of 0.5 is applied during training in all but the last fully connected layers. Additionally, the training images are augmented by computing image transformations such as small shears, rotations and slantings which could be expected for real word images as well.

## IV. EXPERIMENTS

### A. Data sets

The HW-SYNTH data set is a collection of synthetically generated word images [22]. For generating the word images all 26 letters of the Latin alphabet and the numerals 0 to 9 were used. The data set consists of one million word images, each belonging to one of 10,000 word classes obtained from the Hunspell dictionary. The authors use a total of 100 publicly available fonts for randomly generating each word image. The images are rendered by varying the inter character space, stroke width, and the mean foreground and background pixel distributions. Subsequently, the images are smoothed using Gaussian filtering. Every word class is rendered using all letters in the lower case, all letters capitalized, and only the first letter capitalized. The data partition used for training consists of 750,000 images.

The first handwritten data set we use is the George Washington data set. It consists of 4860 words images. They are a part of the 20-page document of correspondence between George Washington and his associates. Since there exists no official training and test partitions for this data set, we use the common approach of performing a four fold cross validation using the splits presented in [13]. The training and test partitions in all four folds contain 3645 and 1215 images respectively. As the documents in the George Washington data set are obtained from the letterbook 2, which is not an original, but a later re-copied volume, it can be assumed that the data set has been produced by a single writer only.

The second handwritten data set is the Esposalles database. It is a multi-writer data set consisting of 32,052 training images and 13,048 test images. The database is an ancient marriage license register written between 1451 and 1905.

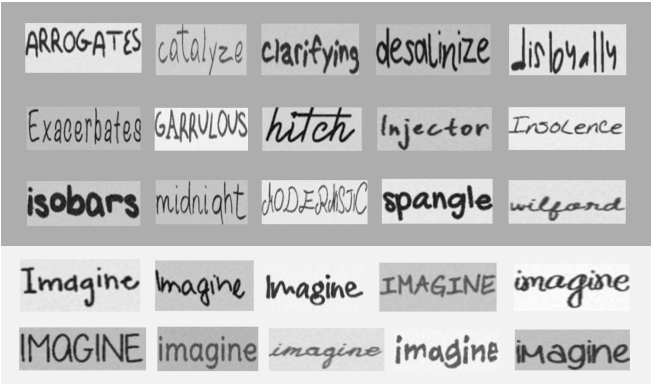

Fig. 1. Top: Examples of random words in the HW-SYNTH data set. Bottom: Examples of a single word from the HW-SYNTH data set showing a variation in fonts and image rendering.

The third handwritten data set is the IAM Handwritten Database written by 657 writers. It consists of 115,320 words and we use the official partition for writer independent text recognition. While the George Washington and Esposalles are historic data sets, the IAM-DB is a contemporary one.

### B. Protocol

The PHOCNet is evaluated for QbE and QbS in the segmentation-based case following the protocol in [13]. The two scenarios we evaluate are training with no manually annotated training data and another with a small amount of annotated training samples. In both cases the PHOCNet is first trained on the HW-SYNTH data set for 80,000 iterations. For performing QbE in the scenario with no added training data, we use each image in the test partitions of the handwritten data sets as a query and predict attribute representations from the CNN for both query and the remaining images which we consider the corpus to run retrieval on. The word images from the corpus are ranked according to the cosine distance of the respective prediction from the PHOCNet and the predicted query representation. We do not use words which appear only once in the test set as queries, since there would be no relevant retrieved image for that query. For QbS, all unique transcriptions for words in the test set are extracted and their PHOC representations are used as queries to rank all corpus representations. For both QbE and QbS we make use of mean Average Precision (mAP) as performance measure.

For the scenario with a limited amount of annotated training data, we use subsets of the respective training partitions of the data sets consisting of an absolute number of word images and finetune the model for 40,000 iterations. In our experiments, we use subsets containing 100, 250, 500, and 1000 word images randomly drawn for each data set. These subsets are augmented in order to increase the number of images using the approach presented in [1]. The performance is evaluated for QbE and QbS using the same protocol as above.

### C. Training Setup

The initialization of parameters of a network plays a crucial role in learning the model. The biases of the PHOCNet are initialized to zero whereas the weights are randomly sampled from a normal distribution with mean 0 and standard deviation $\frac{2}{n}$ [25]. Here, $n$ is the number of parameters in the layer to which the given parameter belongs. We use the exact same meta-parameters for training as were presented in [1]: The PHOCNet is trained using stochastic gradient descent with a batch size of 10. The initial learning rate is set to $10^{-4}$ and is divided by 10 after 70,000 iterations. The momentum is 0.9 and the weight decay is $5 \cdot 10^{-5}$. The second phase of training uses the momentum vector from the previous training phase, and a constant learning rate of $10^{-5}$. The experiments are carried out using a single Nvidia GeForce GTX 1080 GPU.

The word images in the synthetic data set have a fixed size of $48 \times 128$ pixels, whereas those in the handwritten data sets have varying and often larger sizes (depending an the resolution they were scanned with). In order to introduce a suitable variation in image size to our model trained on purely synthetic data, we scale the images in the synthetic data set by a random factor within the interval $[1, 2)$.

### D. Results

The percentage mAP scores for experiments run on the three data sets are listed in Table I. Figure 2 shows the mAP scores for all three data sets over the course of training using only the synthetic data set. Here, especially the experiments run on the Esposalles data set exhibit a strong overfitting behavior. The mAP scores for finetuning the network on the respective data sets are shown in Figure 3. Note that the results in Figure 3 show only the first 1000 iterations, whereas the results in Table I have been obtained after 40,000 iterations.

### E. Discussion

Several inferences can be drawn from our experiments. Firstly, from our results in Table I, we observe that training the PHOCNet on the HW-SYNTH data set (i.e. synthetic data) only does not allow the CNN to perform better than common unsupervised techniques [26]. However, after finetuning the model, its performance for the historic data sets surpasses that of the aforementioned unsupervised techniques using merely 100 images from the handwritten data set. Hence, we are able to outperform unsupervised methods of learning with very limited training data.

Secondly, we are able to reduce the training set size by $86\%$ for GW and almost $98\%$ for IAM and still achieve results competitive to those of an AttributeSVM [13].

The low amount of annotation effort that our method demands can be, further, put into perspective. The ICFHR2016 Handwritten Keyword Spotting Competition [27] states that, for practical applications of keyword spotting, annotating 40 pages of handwritten documents is not a significantly tedious task and that even annotating 154 pages is affordable. With our setup, we are able to drastically reduce this number even further. For example, each page of the George Washington data set consists of approximately 250 words on an average. This implies that, merely 4 pages must be annotated in order to obtain high performance.

TABLE I
RESULTS FOR EXPERIMENTS WITH DIFFERENT AMOUNTS OF TRAINING IMAGES IN MAP [%]

| Method | Training Subset Size | GW QbE | GW QbS | IAM QbE | IAM QbS | Esposalles QbE | Esposalles QbS |
|---|---|---|---|---|---|---|---|
| proposed | 0 | 39.89 | 48.92 | 26.21 | 36.57 | 34.92 | 10.30 |
| | 100 | 83.05 | 86.69 | 38.45 | 56.47 | 89.67 | 71.15 |
| | 250 | 90.76 | 92.39 | 43.78 | 60.90 | 94.06 | 82.43 |
| | 500 | 93.86 | 94.82 | 52.41 | 68.33 | 95.14 | 85.42 |
| | 1000 | 95.74 | 96.59 | 55.39 | 74.09 | 96.27 | 89.18 |
| AttributeSVM [13] | complete | 93.04 | 91.29 | 55.73 | 73.72 | – | – |
| PHOCNet [1] | complete | 96.71 | 92.64 | 72.51 | 82.97 | 97.24 | 93.29 |
| Deep Features [14] | complete | 94.41 | 92.84 | 84.24 | 91.58 | – | – |
| Triplet-CNN [15] | complete | 98.00 | 93.69 | 81.58 | 89.49 | – | – |

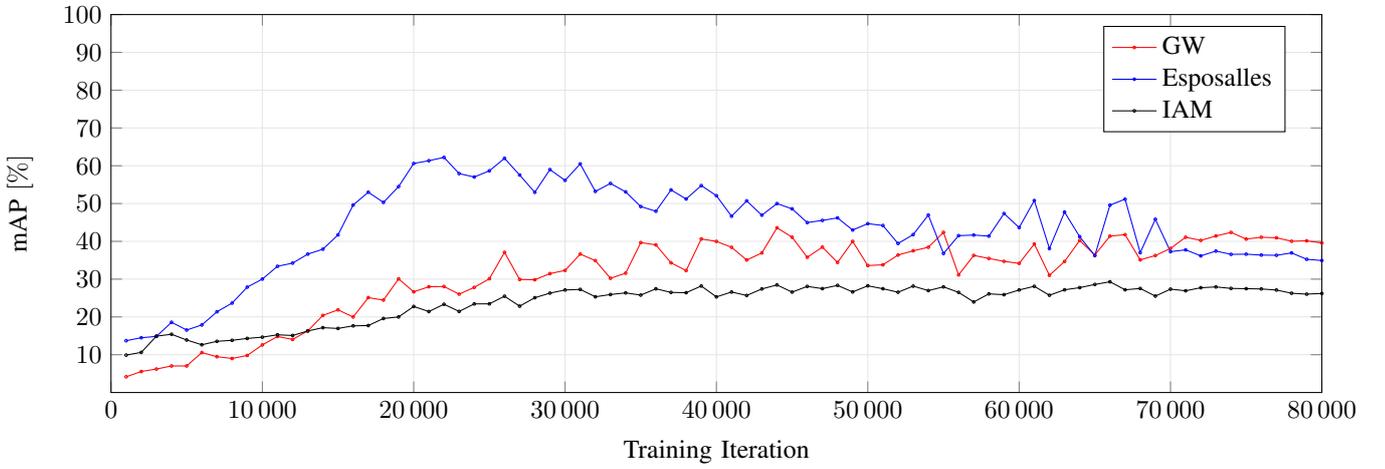

Fig. 2. QbE Mean Average Precision values for evaluating a PHOCNet trained on purely synthetic data on the three respective data sets.

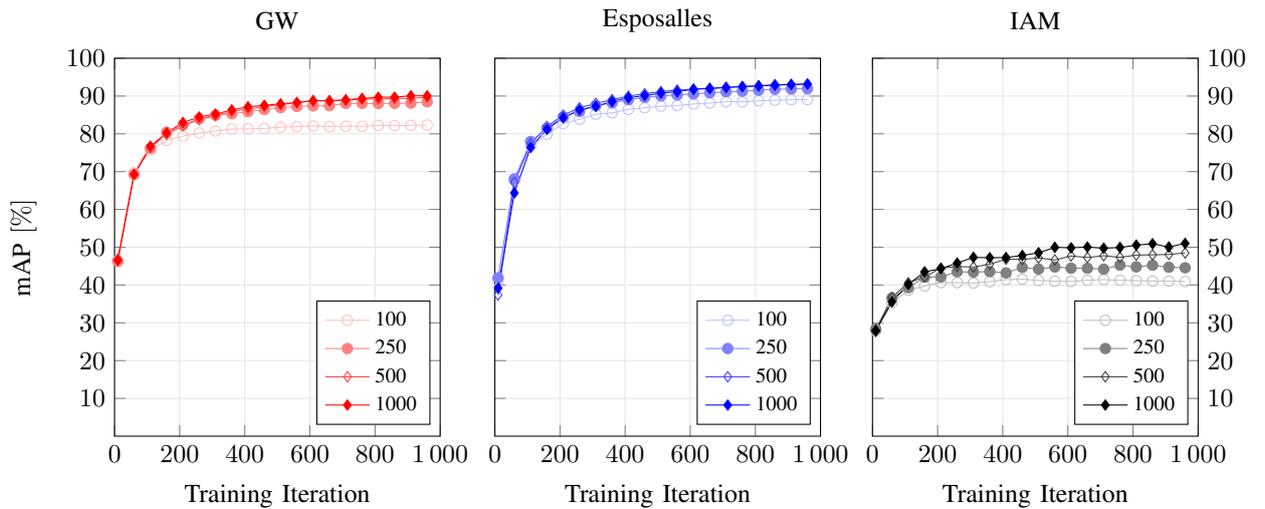

Fig. 3. Mean Average Precision values for the three QbE finetuning experiments. The numbers in the legend indicate how many images were selected for finetuning from the training images.

Another significant feature of our method is the low training time. Although some time is required to train the model on the synthetic data, this phase of training does only need to be carried out once. In fact, if a model pre-trained on the synthetic data is made available, the training time to be considered for a practical application would be only that needed for finetuning the model on the desired handwritten data set. From Figure 3, it can be observed that the network learns the model largely within the first 1000 iterations. The time required to finetune the pre-trained model for 1000 iterations ranges from approximately 7 to 11 minutes, depending on the data set used. This means, that users of the presented system can fit the PHOCNet to their data using only 1000 word images (about four pages of manual annotation needed in the case of the George Washington data set) in 11 minutes or less and obtain competitive word spotting performance.

## V. CONCLUSION

In this work, we presented a system for word spotting using CNNs under weak supervision. We showed empirically, that our system is able to learn the characteristics of unseen handwritten data with minimal manual annotation efforts in a very small time frame while achieving highly competitive results. It is therefore ideally suited for integration into consumer word spotting applications or as initialization for semi-supervised approaches. In those applications, training times can be heavily reduced while not sacrificing performance.

Another aspect of our work is that we were able to show that the PHOCNet is able to achieve competitive performance with highly limited real data, heavily relying on synthetically generated data for training.